\documentclass[final,1p,times,number]{elsarticle}

\pdfoutput=1

\makeatletter
\def\ps@pprintTitle{
 \let\@oddhead\@empty
 \let\@evenhead\@empty
 \def\@oddfoot{\centerline{\thepage}}
 \let\@evenfoot\@oddfoot}
\makeatother

\usepackage{graphicx}
\usepackage{amssymb}
\usepackage{xfrac}
\usepackage{xcolor}
\usepackage{mathtools}
\usepackage{booktabs}
\usepackage{tabularx}
\usepackage{array}
\usepackage{multirow}
\usepackage[colorlinks]{hyperref}

\DeclarePairedDelimiterX{\infdivx}[2]{(}{)}{#1\;\delimsize\|\;#2}
\newcommand{\infdiv}{D_{KL}\infdivx}

\newcolumntype{Y}{>{\centering\arraybackslash}X}
\newcolumntype{P}{>{\centering\arraybackslash}p}

\definecolor{tab_red}{RGB}{194,33,33}
\definecolor{tab_green}{RGB}{33,194,33}

\begin{document}

\begin{frontmatter}

\vspace*{-1cm}

\title{Contextual Encoder-Decoder Network \\ for Visual Saliency Prediction}

\author[add1,add2]{Alexander Kroner\corref{cor1}}
\author[add1,add2]{Mario Senden}
\author[add3]{Kurt Driessens}
\author[add1,add2,add4]{Rainer Goebel}

\address[add1]{Department of Cognitive Neuroscience, Faculty of Psychology and Neuroscience, \\ Maastricht University, Maastricht, The Netherlands}
\address[add2]{Maastricht Brain Imaging Centre, Faculty of Psychology and Neuroscience, \\ Maastricht University, Maastricht, The Netherlands}
\address[add3]{Department of Data Science and Knowledge Engineering, Faculty of Science and Engineering, \\ Maastricht University, Maastricht, The Netherlands}
\address[add4]{Department of Neuroimaging and Neuromodeling, Netherlands Institute for Neuroscience, \\ Royal Netherlands Academy of Arts and Sciences (KNAW), Amsterdam, The Netherlands}

\cortext[cor1]{Corresponding author. \textit{Email address:} \href{mailto:kroner.contact@gmail.com}{kroner.contact@gmail.com}}

\begin{abstract}
Predicting salient regions in natural images requires the detection of objects that are present in a scene. To develop robust representations for this challenging task, high-level visual features at multiple spatial scales must be extracted and augmented with contextual information. However, existing models aimed at explaining human fixation maps do not incorporate such a mechanism explicitly. Here we propose an approach based on a convolutional neural network pre-trained on a large-scale image classification task. The architecture forms an encoder-decoder structure and includes a module with multiple convolutional layers at different dilation rates to capture multi-scale features in parallel. Moreover, we combine the resulting representations with global scene information for accurately predicting visual saliency. Our model achieves competitive and consistent results across multiple evaluation metrics on two public saliency benchmarks and we demonstrate the effectiveness of the suggested approach on five datasets and selected examples. Compared to state of the art approaches, the network is based on a lightweight image classification backbone and hence presents a suitable choice for applications with limited computational resources, such as (virtual) robotic systems, to estimate human fixations across complex natural scenes. Our TensorFlow implementation is openly available at \url{https://github.com/alexanderkroner/saliency}.
\end{abstract}

\end{frontmatter}

\section{Introduction}
Humans demonstrate a remarkable ability to obtain relevant information from complex visual scenes~\cite{jonides1982integrating,irwin1991information}. Overt attention is the mechanism that governs the processing of stimuli by directing gaze towards a spatial location within the visual field~\cite{posner1980orienting}. This sequential selection ensures that the eyes sample prioritized aspects from all available information to reduce the cost of cortical computation~\cite{lennie2003cost}. In addition, only a small central region of the retina, known as the fovea, transforms incoming light into neural responses with high spatial resolution, whereas acuity decreases rapidly towards the periphery~\cite{cowey1974human,berkley1975grating}. Given the limited number of photoreceptors in the eye, this arrangement allows to optimally process visual signals from its environment~\cite{cheung2016emergence}. The function of fixations is thus to resolve the trade-off between coverage and sampling resolution of the whole visual field~\cite{gegenfurtner2016interaction}.

\begin{figure}[t!]
\centering\includegraphics[width=0.75\linewidth]{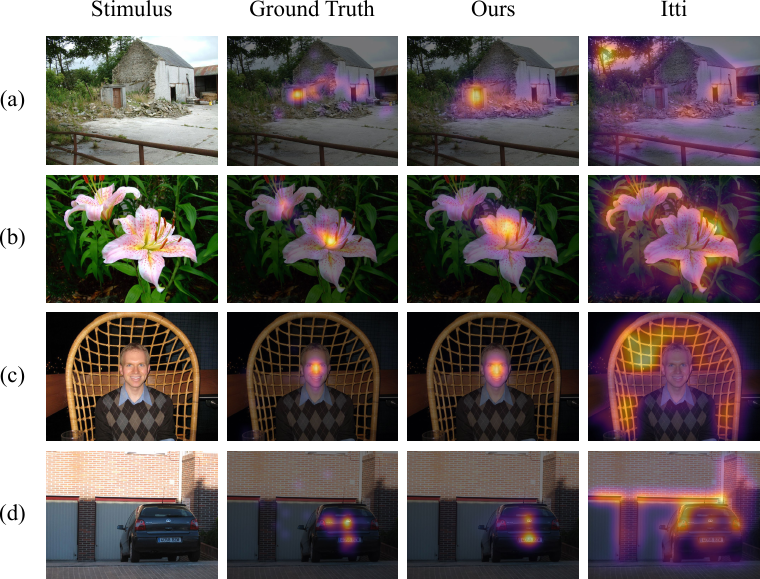}
\caption{A visualization of four natural images with the corresponding empirical fixation maps, our model predictions, and estimated maps based on the work by~\citet{itti1998model}. The network proposed in this study was not trained on the stimuli shown here and thus exhibits its generalization ability to unseen instances. All image examples demonstrate a qualitative agreement of our model with the ground truth data, assigning high saliency to regions that contain semantic information, such as a door (a), flower (b), face (c), or text (d). On the contrary, the approach by~\citet{itti1998model} detected low-level feature contrasts and wrongly predicted high values at object boundaries rather than their center.}
\label{fig:fig1}
\end{figure}

The spatial allocation of attention when viewing natural images is commonly represented in the form of topographic saliency maps that depict which parts of a scene attract fixations reliably. Identifying the underlying properties of these regions would allow us to predict human fixation patterns and gain a deeper understanding of the processes that lead to the observed behavior. In computer vision, this challenging problem has originally been approached using models rooted in \textit{Feature Integration Theory}~\cite{treisman1980feature}. The theory suggests that early visual features must first be registered in parallel before serial shifts of overt attention combine them into unitary object-based representations. This two-stage account of visual processing has emphasized the role of stimulus properties for explaining human gaze. In consequence, the development of feature-driven models has been considered sufficient to enable the prediction of fixation patterns under task-free viewing conditions. \citet{Koch1985ShiftsIS} have introduced the notion of a central saliency map which integrates low-level information and serves as the basis for eye movements. This has resulted in a first model implementation by~\citet{itti1998model} that influenced later work on biologically-inspired architectures.

With the advent of deep neural network solutions for visual tasks such as image classification~\cite{krizhevsky2012imagenet}, saliency modeling has also undergone a paradigm shift from manual feature engineering towards automatic representation learning. In this work, we leveraged the capability of \textit{convolutional neural networks} (CNNs) to extract relevant features from raw images and decode them towards a distribution of saliency across arbitrary scenes. Compared to the seminal work by~\citet{itti1998model}, this approach allows predictions to be based on semantic information instead of low-level feature contrasts (see Figure~\ref{fig:fig1}). This choice was motivated by studies demonstrating the importance of high-level image content for attentional selection in natural images~\cite{einhauser2008objects,nuthmann2010object}.

Furthermore, it is expected that complex representations at multiple spatial scales are necessary for accurate predictions of human fixation patterns. We therefore incorporated a contextual module that samples multi-scale information and augments it with global scene features. The contribution of the contextual module to the overall performance was assessed and final results were compared to previous work on two public saliency benchmarks. We achieved predictive accuracy on unseen test instances at the level of current state of the art approaches, while utilizing a computationally less expensive network backbone with roughly one order of magnitude fewer processing layers. This makes our model suitable for applications in (virtual) robotic environments, as demonstrated by~\citet{bornet2019running}, and we developed a webcam-based interface for saliency prediction in the browser with only moderate hardware requirements (see \url{https://storage.googleapis.com/msi-net/demo/index.html}).

\section{Related Work}

Early approaches towards computational models of visual attention were defined in terms of different theoretical frameworks, including Bayesian~\cite{zhang2008sun} and graph-based formulations~\cite{harel2007graph}. The former was based on the notion of self-information derived from a probability distribution over linear visual features as acquired from natural scenes. The latter framed saliency as the dissimilarity between nodes in a fully-connected directed graph that represents all image locations in a feature map. \citet{hou2007saliency} have instead proposed an approach where images were transformed to the log spectrum and saliency emerged from the spectral residual after removing statistically redundant components. A mechanism inspired more by biological than mathematical principles was first implemented and described in the seminal work by~\citet{itti1998model}. Their model captures center-surround differences at multiple spatial scales with respect to three basic feature channels: color, intensity, and orientation. After normalization of activity levels, the output is fed into a common saliency map depicting local conspicuity in static scenes. This standard cognitive architecture has since been augmented with additional feature channels that capture semantic image content, such as faces and text~\cite{cerf2009faces}.

With the large-scale acquisition of eye tracking measurements under natural viewing conditions, data-driven machine learning techniques became more practicable. \citet{judd2009learning} introduced a model based on support vector machines to estimate fixation densities from a set of low-, mid-, and high-level visual features. While this approach still relied on a hypothesis specifying which image properties would successfully contribute to the prediction of saliency, it marked the beginning of a progression from manual engineering to automatic learning of features. This development has ultimately led to applying deep neural networks with emergent representations for the estimation of human fixation patterns. \citet{vig2014large} were the first to train an ensemble of shallow CNNs to derive saliency maps from natural images in an end-to-end fashion, but failed to capture object information due to limited network depth.

Later attempts addressed that shortcoming by taking advantage of classification architectures pre-trained on the \textit{ImageNet} database~\cite{deng2009imagenet}. This choice was motivated by the finding that features extracted from CNNs generalize well to other visual tasks~\cite{donahue2014decaf}. Consequently, \textit{DeepGaze I}~\cite{kummerer2014deep} and \textit{II}~\cite{kummerer2016deepgaze} employed a pre-trained classification model to read out salient image locations from a small subset of encoding layers. This is similar to the network by~\citet{cornia2016deep} which utilizes the output at three stages of the hierarchy. \citet{oyama2018influence} demonstrated that classification performance of pre-trained architectures strongly correlates with the accuracy of saliency predictions, highlighting the importance of object information. Related approaches also focused on the potential benefits of incorporating activation from both coarse and fine image resolutions~\cite{huang2015salicon}, and recurrent connections to capture long-range spatial dependencies in convolutional feature maps~\cite{Cornia2018PredictingHE,liu2018deep}. Our model explicitly combines semantic representations at multiple spatial scales to include contextual information in the predictive process. For a more complete account of existing saliency architectures, we refer the interested reader to a comprehensive review by~\citet{borji2018saliency}.

\section{Methods}

We propose a new CNN architecture with modules adapted from the semantic segmentation literature to predict fixation density maps of the same image resolution as the input. Our approach is based on a large body of research regarding saliency models that leverage object-specific features and functionally replicate human behavior under free-viewing conditions. In the following sections, we describe our contributions to this challenging task.

\subsection{Architecture}

Image-to-image learning problems require the preservation of spatial features throughout the whole processing stream. As a consequence, our network does not include any fully-connected layers and reduces the number of downsampling operations inherent to classification models. We adapted the popular \textit{VGG16} architecture~\cite{simonyan2014very} as an image encoder by reusing the pre-trained convolutional layers to extract increasingly complex features along its hierarchy. Striding in the last two pooling layers was removed, which yields spatial representations at \sfrac{1}{8} of their original input size. All subsequent convolutional encoding layers were then dilated at a rate of 2 by expanding their kernel, and thereby increased the receptive field to compensate for the higher resolution~\cite{yu2015multi}. This modification still allowed us to initialize the model with pre-trained weights since the number of trainable parameters remained unchanged. Prior work has shown the effectiveness of this approach in the context of saliency prediction problems~\cite{Cornia2018PredictingHE,liu2018deep}.

\begin{figure}[t!]
\centering\includegraphics[width=\linewidth]{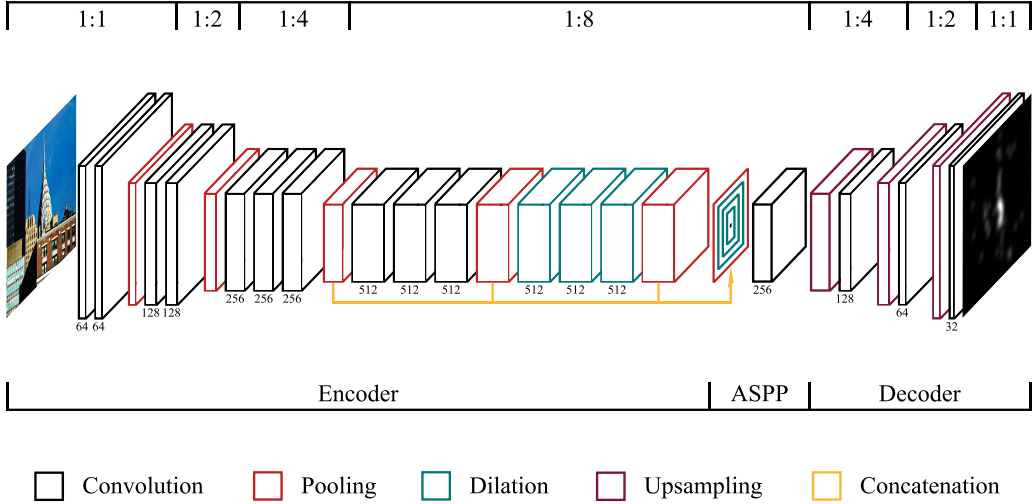}
\caption{An illustration of the modules that constitute our encoder-decoder architecture. The VGG16 backbone was modified to account for the requirements of dense prediction tasks by omitting feature downsampling in the last two max-pooling layers. Multi-level activations were then forwarded to the ASPP module, which captured information at different spatial scales in parallel. Finally, the input image dimensions were restored via the decoder network. Subscripts beneath convolutional layers denote the corresponding number of feature maps.} \label{fig:fig2}
\end{figure}

For related visual tasks such as semantic segmentation, information distributed over convolutional layers at different levels of the hierarchy can aid the preservation of fine spatial details~\cite{hariharan2015hypercolumns,long2015fully}. The prediction of fixation density maps does not require accurate class boundaries but still benefits from combined mid- to high-level feature responses~\cite{kummerer2014deep,kummerer2016deepgaze,cornia2016deep}. Hence, we adapted the multi-level design proposed by~\citet{cornia2016deep} and concatenated the output from layers 10, 14, and 18 into a common tensor with 1,280 activation maps.

This representation constitutes the input to an \textit{Atrous Spatial Pyramid Pooling} (ASPP) module~\cite{chen2018deeplab}. It utilizes several convolutional layers with different dilation factors in parallel to capture multi-scale image information. Additionally, we incorporated scene content via global average pooling over the final encoder output, as motivated by the study of~\citet{torralba2006contextual} who stated that contextual information plays an important role for the allocation of attention. Our implementation of the ASPP architecture thus closely follows the modifications proposed by~\citet{chen2017rethinking}. These authors augmented multi-scale information with global context and demonstrated performance improvements on semantic segmentation tasks.

\newpage

In this work, we laid out three convolutional layers with kernel sizes of $3\times3$ and dilation rates of 4, 8, and 12 in parallel, together with a $1\times1$ convolutional layer that could not learn new spatial dependencies but nonlinearly combined existing feature maps. Image-level context was represented as the output after global average pooling (i.e. after averaging the entries of a tensor across both spatial dimensions to a single value) and then brought to the same resolution as all other representations via bilinear upsampling, followed by another point-wise convolutional operation. Each of the five branches in the module contains 256 filters, which resulted in an aggregated tensor of 1,280 feature maps. Finally, the combined output was forwarded to a $1\times1$ convolutional layer with 256 channels that contained the resulting multi-scale responses.

To restore the original image resolution, extracted features were processed by a series of convolutional and upsampling layers. Previous work on saliency prediction has commonly utilized bilinear interpolation for that task~\cite{Cornia2018PredictingHE,liu2018deep}, but we argue that a carefully chosen decoder architecture, similar to the model by~\citet{pan2017salgan}, results in better approximations. Here we employed three upsampling blocks consisting of a bilinear scaling operation, which doubled the number of rows and columns, and a subsequent convolutional layer with kernel size $3\times3$. This setup has previously been shown to prevent checkerboard artifacts in the upsampled image space in contrast to deconvolution~\cite{odena2016deconvolution}. Besides an increase of resolution throughout the decoder, the amount of channels was halved in each block to yield 32 feature maps. Our last network layer transformed activations into a continuous saliency distribution by applying a final $3\times3$ convolution. The outputs of all but the last linear layer were modified via rectified linear units. Figure~\ref{fig:fig2} visualizes the overall architecture design as described in this section.

\subsection{Training}

Weight values from the ASPP module and decoder were initialized according to the \textit{Xavier} method by~\citet{glorot2010understanding}. It specifies parameter values as samples drawn from a uniform distribution with zero mean and a variance depending on the total number of incoming and outgoing connections. Such initialization schemes are demonstrably important for training deep neural networks successfully from scratch~\cite{sutskever2013importance}. The encoding layers were based on the VGG16 architecture pre-trained on both \textit{ImageNet}~\cite{deng2009imagenet} and \textit{Places2}~\cite{zhou2017places} data towards object and scene classification respectively.

We normalized the model output such that all values are non-negative with unit sum. The estimation of saliency maps can hence be regarded as a \textit{probability distribution prediction} task as formulated by~\citet{jetley2016end}. To determine the difference between an estimated and a target distribution, the \textit{Kullback-Leibler} (KL) divergence is an appropriate measure rooted in information theory to quantify the statistical distance $D$. This can be defined as follows:

\begin{equation}\label{eq:kld}
\infdiv{P}{Q} = \sum \limits_{i} Q_i \ln (\epsilon + \frac{Q_i}{\epsilon + P_i})
\end{equation}

Here, $Q$ represents the target distribution, $P$ its approximation, $i$ each pixel index, and $\epsilon$ a regularization constant. Equation \eqref{eq:kld} served as the loss function which was gradually minimized via the \textit{Adam} optimization algorithm~\cite{kingma2014adam}. We defined an upper learning rate of $10^{-6}$ and modified the weights in an online fashion due to a general inefficiency of batch training according to~\citet{wilson2003general}. Based on this general setup, we trained our network for 10 epochs and used the best-performing checkpoint for inference.

\section{Experiments}

The proposed encoder-decoder model was evaluated on five publicly available eye tracking datasets that yielded qualitative and quantitative results. First, we provide a brief description of the images and empirical measurements utilized in this study. Second, the different metrics commonly used to assess the predictive performance of saliency models are summarized. Finally, we report the contribution of our architecture design choices and benchmark the overall results against baselines and related work in computer vision.

\subsection{Datasets}

A prerequisite for the successful application of deep learning techniques is a wealth of annotated data. Fortunately, the growing interest in developing and evaluating fixation models has lead to the release of large-scale eye tracking datasets such as \textit{MIT1003}~\cite{judd2009learning}, \textit{CAT2000}~\cite{borji2015cat2000}, \textit{DUT-OMRON}~\cite{yang2013saliency}, \textit{PASCAL-S}~\cite{li2014secrets}, and \textit{OSIE}~\cite{xu2014predicting}. The costly acquisition of measurements, however, is a limiting factor for the number of stimuli. New data collection methodologies have emerged that leverage webcam-based eye movements~\cite{xu2015turkergaze} or mouse movements~\cite{jiang2015salicon} instead via crowdsourcing platforms. The latter approach resulted in the \textit{SALICON} dataset, which consists of 10,000 training and 5,000 validation instances serving as a proxy for empirical gaze measurements. Due to its large size, we first trained our model on SALICON before fine-tuning the learned weights towards fixation predictions on either of the other datasets with the same optimization parameters. This widely adopted procedure has been shown to improve the accuracy of eye movement estimations despite some disagreement between data originating from gaze and mouse tracking experiments~\cite{tavakoli2017saliency}.

The images presented during the acquisition of saliency maps in all aforementioned datasets are largely based on natural scenes. Stimuli of CAT2000 additionally fall into predefined categories such as \textit{Action}, \textit{Fractal}, \textit{Object}, or \textit{Social}. Together with the corresponding fixation patterns, they constituted the input and desired output to our network architecture. In detail, we rescaled and padded all images from the SALICON and OSIE datasets to $240\times320$ pixels, the MIT1003, DUT-OMRON, and PASCAL-S datasets to $360\times360$ pixels, and the CAT2000 dataset to $216\times384$ pixels, such that the original aspect ratios were preserved. For the latter five eye tracking sets we defined 80\% of the samples as training data and the remainder as validation examples with a minimum of 200 instances. The correct saliency distributions on test set images of MIT1003 and CAT2000 are held out and predictions must hence be submitted online for evaluation.

\subsection{Metrics}

Various measures are used in the literature and by benchmarks to evaluate the performance of fixation models. In practice, results are typically reported for all of them to include different notions about saliency and allow a fair comparison of model predictions~\cite{kummerer2018saliency,riche2013saliency}. A set of nine metrics is commonly selected: \textit{Kullback-Leibler divergence} (KLD), \textit{Pearson's correlation coefficient} (CC), \textit{histogram intersection} (SIM), \textit{Earth Mover's distance} (EMD), \textit{information gain} (IG), \textit{normalized scanpath saliency} (NSS), and three variants of \textit{area under ROC curve} (AUC-Judd, AUC-Borji, shuffled AUC). The former four are location-based metrics, which require ground truth maps as binary fixation matrices. By contrast, the remaining metrics quantify saliency approximations after convolving gaze locations with a Gaussian kernel and representing the target output as a probability distribution. We refer readers to an overview by~\citet{bylinskii2018different} for more information regarding the implementation details and properties of the stated measures.

In this work, we adopted KLD as an objective function and produced fixation density maps as output from our proposed network. This training setup is particularly sensitive to false negative predictions and thus the appropriate choice for applications aimed at salient target detection~\cite{bylinskii2018different}. Defining the problem of saliency prediction in a probabilistic framework also enables fair model ranking on public benchmarks for the MIT1003, CAT2000, and SALICON datasets~\cite{kummerer2018saliency}. As a consequence, we evaluated our estimated gaze distributions without applying any metric-specific postprocessing methods.

\subsection{Results}

A quantitative comparison of results on independent test datasets was carried out to characterize how well our proposed network generalizes to unseen images. Here, we were mainly interested in estimating human eye movements and regarded mouse tracking measurements merely as a substitute for attention. The final outcome for the 2017 release of the SALICON dataset is therefore not reported in this work but our model results can be viewed on the public leaderboard\footnote{\url{https://competitions.codalab.org/competitions/17136}} under the user name \texttt{akroner}.

\begin{table}[t!]
\centering
\scalebox{0.76}{
\begin{tabularx}{17.6cm}{P{4.12cm} Y Y Y Y Y Y Y Y}
    \toprule
    & \textit{\scalebox{0.9}{AUC-J}} \raisebox{1.3pt}{\scalebox{0.64}{$\uparrow$}} & 
    \textit{\scalebox{0.9}{SIM}} \raisebox{1.3pt}{\scalebox{0.64}{$\uparrow$}} & 
    \textit{\scalebox{0.9}{EMD}} \raisebox{1.3pt}{\scalebox{0.64}{$\downarrow$}} & 
    \textit{\scalebox{0.9}{AUC-B}} \raisebox{1.3pt}{\scalebox{0.64}{$\uparrow$}} & 
    \textit{\scalebox{0.9}{sAUC}} \raisebox{1.3pt}{\scalebox{0.64}{$\uparrow$}} & 
    \textit{\scalebox{0.9}{CC}} \raisebox{1.3pt}{\scalebox{0.64}{$\uparrow$}} & 
    \textit{\scalebox{0.9}{NSS}} \raisebox{1.3pt}{\scalebox{0.64}{$\uparrow$}} & 
    \textit{\scalebox{0.9}{KLD}} \raisebox{1.3pt}{\scalebox{0.64}{$\downarrow$}} \\
    \midrule
    DenseSal \citep{oyama2018influence} & 0.87 & 0.67 & 1.99 & 0.81 & 0.72 & 0.79 & 2.25 & \textbf{0.48} \\
    \addlinespace[0.10em]
    DPNSal \citep{oyama2018influence} & 0.87 & \textbf{0.69} & 2.05 & 0.80 & \textbf{0.74} & \textbf{0.82} & 2.41 & 0.91 \\
    \addlinespace[0.10em]
    SALICON \citep{huang2015salicon}$^\dagger$ & 0.87 & 0.60 & 2.62 & 0.85 & \textbf{0.74} & 0.74 & 2.12 & 0.54 \\
    \addlinespace[0.10em]
    DSCLRCN \citep{liu2018deep} & 0.87 & 0.68 & 2.17 & 0.79 & 0.72 & 0.80 & 2.35 & 0.95 \\
    \addlinespace[0.10em]
    DeepFix \citep{kruthiventi2017deepfix}$^\dagger$ & 0.87 & 0.67 & 2.04 & 0.80 & 0.71 & 0.78 & 2.26 & 0.63 \\
    \addlinespace[0.10em]
    EML-NET \citep{jia2018eml} & \textbf{0.88} & 0.68 & \textbf{1.84} & 0.77 & 0.70 & 0.79 & \textbf{2.47} & 0.84 \\
    \addlinespace[0.10em]
    DeepGaze II \citep{kummerer2016deepgaze} & \textbf{0.88} & 0.46 & 3.98 & \textbf{0.86} & 0.72 & 0.52 & 1.29 & 0.96 \\
    \addlinespace[0.10em]
    SAM-VGG \citep{Cornia2018PredictingHE}$^\dagger$ & 0.87 & 0.67 & 2.14 & 0.78 & 0.71 & 0.77 & 2.30 & 1.13 \\
    \addlinespace[0.10em]
    ML-Net \citep{cornia2016deep}$^\dagger$ & 0.85 & 0.59 & 2.63 & 0.75 & 0.70 & 0.67 & 2.05 & 1.10 \\
    \addlinespace[0.10em]
    SAM-ResNet \citep{Cornia2018PredictingHE} & 0.87 & 0.68 & 2.15 & 0.78 & 0.70 & 0.78 & 2.34 & 1.27 \\
    \addlinespace[0.10em]
    DeepGaze I \citep{kummerer2014deep} & 0.84 & 0.39 & 4.97 & 0.83 & 0.66 & 0.48 & 1.22 & 1.23 \\
    \midrule
    Judd \citep{judd2009learning} & 0.81 & 0.42 & 4.45 & 0.80 & 0.60 & 0.47 & 1.18 & 1.12 \\
    \addlinespace[0.10em]
    eDN \citep{vig2014large} & 0.82 & 0.41 & 4.56 & 0.81 & 0.62 & 0.45 & 1.14 & 1.14 \\
    \midrule
    GBVS \citep{harel2007graph} & 0.81 & 0.48 & 3.51 & 0.80 & 0.63 & 0.48 & 1.24 & 0.87 \\
    \addlinespace[0.10em]
    Itti \citep{itti1998model} & 0.75 & 0.44 & 4.26 & 0.74 & 0.63 & 0.37 & 0.97 & 1.03 \\
    \addlinespace[0.10em]
    SUN \citep{zhang2008sun} & 0.67 & 0.38 & 5.10 & 0.66 & 0.61 & 0.25 & 0.68 & 1.27 \\
    \midrule
    \textbf{Ours}$^\dagger$ & 0.87 & 0.68 & 1.99 & 0.82 & 0.72 & 0.79 & 2.27 & 0.66 \\
    \bottomrule
\end{tabularx}}
\caption{Quantitative results of our model for the MIT300 test set in the context of prior work. The first line separates deep learning approaches with architectures pre-trained on image classification (the superscript $^\dagger$ represents models with a VGG16 backbone) from shallow networks and other machine learning methods. Entries between the second and the third line are models based on theoretical considerations and define a baseline rather than competitive performance. Arrows indicate whether the metrics assess similarity \protect\raisebox{1.0pt}{\protect\scalebox{0.70}{$\uparrow$}} or dissimilarity \protect\raisebox{1.0pt}{\protect\scalebox{0.70}{$\downarrow$}} between predictions and targets. The best results are marked in bold and models are sorted in descending order of their cumulative rank across a subset of weakly correlated evaluation measures within each group.}
\label{tab:tab1} 
\end{table}
\begin{table}[t!]
\centering
\scalebox{0.76}{
\begin{tabularx}{17.6cm}{P{4.12cm} Y Y Y Y Y Y Y Y}
    \toprule
    & \textit{\scalebox{0.9}{AUC-J}} \raisebox{1.3pt}{\scalebox{0.64}{$\uparrow$}} & 
    \textit{\scalebox{0.9}{SIM}} \raisebox{1.3pt}{\scalebox{0.64}{$\uparrow$}} & 
    \textit{\scalebox{0.9}{EMD}} \raisebox{1.3pt}{\scalebox{0.64}{$\downarrow$}} & 
    \textit{\scalebox{0.9}{AUC-B}} \raisebox{1.3pt}{\scalebox{0.64}{$\uparrow$}} & 
    \textit{\scalebox{0.9}{sAUC}} \raisebox{1.3pt}{\scalebox{0.64}{$\uparrow$}} & 
    \textit{\scalebox{0.9}{CC}} \raisebox{1.3pt}{\scalebox{0.64}{$\uparrow$}} & 
    \textit{\scalebox{0.9}{NSS}} \raisebox{1.3pt}{\scalebox{0.64}{$\uparrow$}} & 
    \textit{\scalebox{0.9}{KLD}} \raisebox{1.3pt}{\scalebox{0.64}{$\downarrow$}} \\
    \midrule
    SAM-VGG \citep{Cornia2018PredictingHE}$^\dagger$ & \textbf{0.88} & 0.76 & 1.07 & 0.79 & 0.58 & \textbf{0.89} & \textbf{2.38} & 0.54 \\
    \addlinespace[0.10em]
    SAM-ResNet \citep{Cornia2018PredictingHE} & \textbf{0.88} & \textbf{0.77} & \textbf{1.04} & 0.80 & 0.58 & \textbf{0.89} & \textbf{2.38} & 0.56 \\
    \addlinespace[0.10em]
    DeepFix \citep{kruthiventi2017deepfix}$^\dagger$ & 0.87 & 0.74 & 1.15 & 0.81 & 0.58 & 0.87 & 2.28 & 0.37 \\
    \addlinespace[0.10em]
    EML-NET \citep{jia2018eml} & 0.87 & 0.75 & 1.05 & 0.79 & \textbf{0.59} & 0.88 & \textbf{2.38} & 0.96 \\
    \midrule
    Judd \citep{judd2009learning} & 0.84 & 0.46 & 3.60 & \textbf{0.84} & 0.56 & 0.54 & 1.30 & 0.94 \\
    \addlinespace[0.10em]
    eDN \citep{vig2014large} & 0.85 & 0.52 & 2.64 & \textbf{0.84} & 0.55 & 0.54 & 1.30 & 0.97 \\
    \midrule
    Itti \citep{itti1998model} & 0.77 & 0.48 & 3.44 & 0.76 & \textbf{0.59} & 0.42 & 1.06 & 0.92 \\
    \addlinespace[0.10em]
    GBVS \citep{harel2007graph} & 0.80 & 0.51 & 2.99 & 0.79 & 0.58 & 0.50 & 1.23 & 0.80 \\
    \addlinespace[0.10em]
    SUN \citep{zhang2008sun} & 0.70 & 0.43 & 3.42 & 0.69 & 0.57 & 0.30 & 0.77 & 2.22 \\
    \midrule
    \textbf{Ours}$^\dagger$ & \textbf{0.88} & 0.75 & 1.07 & 0.82 & \textbf{0.59} & 0.87 & 2.30 & \textbf{0.36} \\
    \bottomrule
\end{tabularx}}
\caption{Quantitative results of our model for the CAT2000 test set in the context of prior work. The first line separates deep learning approaches with architectures pre-trained on image classification (the superscript $^\dagger$ represents models with a VGG16 backbone) from shallow networks and other machine learning methods. Entries between the second and third lines are models based on theoretical considerations and define a baseline rather than competitive performance. Arrows indicate whether the metrics assess similarity \protect\raisebox{1.0pt}{\protect\scalebox{0.70}{$\uparrow$}} or dissimilarity \protect\raisebox{1.0pt}{\protect\scalebox{0.70}{$\downarrow$}} between predictions and targets. The best results are marked in bold and models are sorted in descending order of their cumulative rank across a subset of weakly correlated evaluation measures within each group.}
\label{tab:tab2} 
\end{table}

To assess the predictive performance for eye tracking measurements, the MIT saliency benchmark~\cite{bylinskii2015saliency} is commonly used to compare model results on two test datasets with respect to prior work. Final scores can then be submitted on a public leaderboard to allow fair model ranking on eight evaluation metrics. Table~\ref{tab:tab1} summarizes our results on the test dataset of MIT1003, namely \textit{MIT300}~\cite{judd2012benchmark}, in the context of previous approaches. The evaluation shows that our model only marginally failed to achieve state-of-the-art performance on any of the individual metrics. When computing the cumulative rank (i.e. the sum of ranks according to the standard competition ranking procedure) on a subset of weakly correlated measures (sAUC, CC, KLD)~\cite{riche2013saliency,bylinskii2018different}, we ranked third behind the two architectures \textit{DenseSal} and \textit{DPNSal} from~\citet{oyama2018influence}. However, their approaches were based on a pre-trained \textit{Densely Connected Convolutional Network} with 161 layers~\cite{huang2017densely} and \textit{Dual Path Network} with 131 layers~\cite{chen2017dual} respectively, both of which are computationally far more expensive than the VGG16 model used in this work (see Table~5 by~\citet{oyama2018influence} for a comparison of the computational efficiency). Furthermore, DenseSal and DPNSal implemented a multi-path design where two images of different resolutions are simultaneously fed to the network, which substantially reduces the execution speed compared to single-stream architectures. Among all entries of the MIT300 benchmark with a VGG16 backbone~\cite{cornia2016deep,huang2015salicon,Cornia2018PredictingHE,kruthiventi2017deepfix}, our model clearly achieved the highest performance.

\begin{table}[t!]
\centering
\scalebox{0.76}{
\begin{tabularx}{7.0cm}{Y Y}
    \toprule
    & \textit{Parameters} \\
    \midrule
    DeepGaze I \citep{kummerer2014deep} & 3,750,913 \\
    \addlinespace[0.10em]
    ML-Net \citep{cornia2016deep}$^\dagger$ & 15,452,145 \\
    \addlinespace[0.10em]
    DeepGaze II \citep{kummerer2016deepgaze} & 20,065,973 \\
    \addlinespace[0.10em]
    DenseSal \citep{oyama2018influence} & 26,474,209 \\
    \addlinespace[0.10em]
    SALICON \citep{huang2015salicon}$^\dagger$ & 29,429,889 \\
    \addlinespace[0.10em]
    DSCLRCN \citep{liu2018deep} & 30,338,437 \\
    \addlinespace[0.10em]
    DeepFix \citep{kruthiventi2017deepfix}$^\dagger$ & 35,455,617 \\
    \addlinespace[0.10em]
    SAM-VGG \citep{Cornia2018PredictingHE}$^\dagger$ & 51,835,841 \\
    \addlinespace[0.10em]
    SAM-ResNet \citep{Cornia2018PredictingHE} & 70,093,441 \\
    \addlinespace[0.10em]
    DPNSal \citep{oyama2018influence} & 76,536,513 \\
    \addlinespace[0.10em]
    EML-NET \citep{jia2018eml} & 84,728,569 \\
    \midrule
    \textbf{Ours} & 24,934,209 \\
    \bottomrule
\end{tabularx}}
\caption{The number of trainable parameters for all deep learning models listed in Table~\ref{tab:tab1} that are competing in the MIT300 saliency benchmark. Entries of prior work are sorted according to increasing network complexity and the superscript $^\dagger$ represents pre-trained models with a VGG16 backbone.}
\label{tab:tab3} 
\end{table}
\begin{table}[t!]
\centering
\scalebox{0.76}{
\begin{tabular}{P{3.0cm} P{2.2cm} P{2.0cm} P{1.7cm} P{2.6cm}}
    \toprule
     & \textit{Sizes} & \textit{Speed} & \textit{Memory} & \textit{Computations} \\
    \midrule
    MIT1003 & $360\times360$ px & 43 FPS & 194 MB & 75 GFLOPS \\
    \addlinespace[0.10em]
    CAT2000 & $216\times384$ px & 56 FPS & 175 MB & 48 GFLOPS \\
    \addlinespace[0.10em]
    DUT-OMRON & $360\times360$ px & 43 FPS & 194 MB & 75 GFLOPS \\
    \addlinespace[0.10em]
    PASCAL-S & $360\times360$ px & 43 FPS & 194 MB & 75 GFLOPS \\
    \addlinespace[0.10em]
    OSIE & $240\times320$ px & 58 FPS & 173 MB & 44 GFLOPS \\
    \bottomrule
\end{tabular}}
\caption{The results after evaluating our model with respect to its computational efficiency. We tested five versions trained on different eye tracking datasets, each receiving input images of their preferred sizes in \textit{pixels}~(px). After running each network on 10,000 test set instances from the ImageNet database for 10 times, we averaged the inference speed and described the results in \textit{frames per second}~(FPS). All settings demonstrated consistent outcomes with a standard deviation of less than 1 FPS. The minimal GPU memory utilization was measured with TensorFlow in \textit{megabytes}~(MB) and included the requirements for initializing a testing session. Finally, we estimated the \textit{floating point operations per second}~(FLOPS) at a scale of 9 orders of magnitude.}
\label{tab:tab4} 
\end{table}

\begin{table}[t!]
\centering
\scalebox{0.76}{
\begin{tabular}{P{1.5cm} P{3.5cm}}
    \toprule
    \multicolumn{2}{c}{\textit{Hardware specifications}} \\
    \midrule
    GPU & NVIDIA TITAN Xp \\
    \addlinespace[0.10em]
    CPU & Intel Xeon E5-1650 \\
    \addlinespace[0.10em]
    RAM & 32 GB DDR4 \\
    \addlinespace[0.10em]
    HDD & 256 GB SSD \\
    \bottomrule
    \end{tabular}
    \quad\quad\quad\quad
    \begin{tabular}{P{2.9cm} P{2.1cm}}
    \toprule
    \multicolumn{2}{c}{\textit{Software specifications}} \\
    \midrule
    TensorFlow & 1.14.0 \\
    \addlinespace[0.10em]
    CUDA & 10.0 \\
    \addlinespace[0.10em]
    cuDNN & 7.5 \\
    \addlinespace[0.10em]
    GPU driver & 418.74 \\
    \bottomrule
\end{tabular}}
\caption{Details regarding the hardware and software specifications used throughout our evaluation of computational efficiency. The system ran under the Debian 9 operating system and we minimized usage of the computer during the experiments to avoid interference with measurements of inference speed.}
\label{tab:tab5}
\end{table}

\newpage

We further evaluated the model complexity of all relevant deep learning approaches listed in Table~\ref{tab:tab1}. The number of trainable parameters was computed based on either the official code repository or a replication of the described architectures. In case a reimplementation was not possible, we faithfully estimated a lower bound given the pre-trained classification network. Table~\ref{tab:tab3} summarizes the findings and shows that our model compares favorably to the best-performing approaches. While the number of parameters provides an indication about the computational efficiency of an algorithm, more measures are needed. Therefore, we recorded the inference speed and GPU memory consumption of our model and calculated the number of computations (see Table~\ref{tab:tab4}) for our given hardware and software specifications (see Table~\ref{tab:tab5}). The results highlight that our approach achieves fast inference speed combined with a low GPU memory footprint, and thus enables applications to systems constrained by computational resources.

Table~\ref{tab:tab2} demonstrates that we obtained state-of-the-art scores for the CAT2000 test dataset regarding the AUC-J, sAUC, and KLD evaluation metrics, and competitive results on the remaining measures. The cumulative rank (as computed above) suggests that our model outperformed all previous approaches, including the ones based on a pre-trained VGG16 classification network~\citep{Cornia2018PredictingHE,kruthiventi2017deepfix}. Our final evaluation results for both the MIT300 and CAT2000 datasets can be viewed on the MIT saliency benchmark under the model name \texttt{MSI-Net}, representing our multi-scale information network. Qualitatively, the proposed architecture successfully captures semantically meaningful image features such as faces and text towards the prediction of saliency, as can be seen in Figure~\ref{fig:fig1}. Unfortunately, a visual comparison with the results from prior work was not possible since most models are not openly available.

\begin{table}[t!]
\centering
\scalebox{0.76}{
\begin{tabularx}{17.6cm}{P{2.25cm} P{1.7cm} P{0.05cm} Y Y Y Y Y Y Y Y}
    \toprule
    & & & \textit{\scalebox{0.9}{AUC-J}} \raisebox{1.3pt}{\scalebox{0.64}{$\uparrow$}} & 
    \textit{\scalebox{0.9}{SIM}} \raisebox{1.3pt}{\scalebox{0.64}{$\uparrow$}} & 
    \textit{\scalebox{0.9}{EMD}} \raisebox{1.3pt}{\scalebox{0.64}{$\downarrow$}} & 
    \textit{\scalebox{0.9}{AUC-B}} \raisebox{1.3pt}{\scalebox{0.64}{$\uparrow$}} & 
    \textit{\scalebox{0.9}{sAUC}} \raisebox{1.3pt}{\scalebox{0.64}{$\uparrow$}} & 
    \textit{\scalebox{0.9}{CC}} \raisebox{1.3pt}{\scalebox{0.64}{$\uparrow$}} & 
    \textit{\scalebox{0.9}{NSS}} \raisebox{1.3pt}{\scalebox{0.64}{$\uparrow$}} & 
    \textit{\scalebox{0.9}{KLD}} \raisebox{1.3pt}{\scalebox{0.64}{$\downarrow$}} \\
    \midrule
    \multirow{4}{*}{\scalebox{1.10}{MIT1003}} & \multirow{2}{*}{\raisebox{1.0pt}{$\oplus$}ASPP} & $\mu$ & \hspace{0.4em}\textbf{0.899}* & \hspace{0.4em}\textbf{0.602}* & \hspace{0.4em}\textbf{2.430}* & \hspace{0.4em}\textbf{0.832}* & 0.713 & \hspace{0.4em}\textbf{0.741}* & \hspace{0.4em}\textbf{2.663}* & \hspace{0.4em}\textbf{0.818}* \\
    & & $\sigma$ & 0.001 & 0.004 & 0.033 & 0.005 & 0.003 & 0.003 & 0.012 & 0.051 \\
    \addlinespace[0.15em]
    & \multirow{2}{*}{\raisebox{1.0pt}{$\ominus$}ASPP} & $\mu$ & 0.890 & 0.573 & 2.645 & 0.827 & \textbf{0.720} & 0.700 & 2.540 & 0.867 \\
    & & $\sigma$ & 0.001 & 0.005 & 0.033 & 0.006 & 0.003 & 0.004 & 0.016 & 0.042 \\
    \midrule
    \multirow{4}{*}{\scalebox{1.10}{CAT2000}} & \multirow{2}{*}{\raisebox{1.0pt}{$\oplus$}ASPP} & $\mu$ & \hspace{0.4em}\textbf{0.882}* & \hspace{0.4em}\textbf{0.734}* & \hspace{0.4em}\textbf{2.553}* & 0.812 & 0.582 & \hspace{0.4em}\textbf{0.854}* & \hspace{0.4em}\textbf{2.359}* & \hspace{0.4em}\textbf{0.430}* \\
    & & $\sigma$ & 0.000 & 0.002 & 0.025 & 0.005 & 0.003 & 0.003 & 0.007 & 0.010 \\
    \addlinespace[0.15em]
    & \multirow{2}{*}{\raisebox{1.0pt}{$\ominus$}ASPP} & $\mu$ & 0.873 & 0.683 & 2.975 & \textbf{0.824} & \textbf{0.591} & 0.770 & 2.092 & 0.501 \\
    & & $\sigma$ & 0.000 & 0.002 & 0.033 & 0.004 & 0.002 & 0.002 & 0.006 & 0.013 \\
    \midrule
    \multirow{4}{*}{\scalebox{1.10}{DUT-OMRON}} & \multirow{2}{*}{\raisebox{1.0pt}{$\oplus$}ASPP} & $\mu$ & \hspace{0.4em}\textbf{0.921}* & \hspace{0.4em}\textbf{0.649}* & \hspace{0.4em}\textbf{1.155}* & \hspace{0.4em}\textbf{0.864}* & 0.775 & \hspace{0.4em}\textbf{0.776}* & \hspace{0.4em}\textbf{2.873}* & \hspace{0.4em}\textbf{0.634}* \\
    & & $\sigma$ & 0.001 & 0.002 & 0.016 & 0.005 & 0.003 & 0.002 & 0.008 & 0.023 \\
    \addlinespace[0.15em]
    & \multirow{2}{*}{\raisebox{1.0pt}{$\ominus$}ASPP} & $\mu$ & 0.915 & 0.627 & 1.237 & 0.855 & \textbf{0.777} & 0.748 & 2.789 & 0.695 \\
    & & $\sigma$ & 0.001 & 0.003 & 0.022 & 0.004 & 0.003 & 0.002 & 0.011 & 0.031 \\
    \midrule
    \multirow{4}{*}{\scalebox{1.10}{PASCAL-S}} & \multirow{2}{*}{\raisebox{1.0pt}{$\oplus$}ASPP} & $\mu$ & \hspace{0.4em}\textbf{0.914}* & \hspace{0.4em}\textbf{0.667}* & \hspace{0.4em}\textbf{1.015}* & \hspace{0.4em}\textbf{0.853}* & 0.701 & \hspace{0.4em}\textbf{0.818}* & \hspace{0.4em}\textbf{2.610}* & \hspace{0.4em}\textbf{0.645}* \\
    & & $\sigma$ & 0.000 & 0.003 & 0.010 & 0.004 & 0.003 & 0.002 & 0.008 & 0.044 \\
    \addlinespace[0.15em]
    & \multirow{2}{*}{\raisebox{1.0pt}{$\ominus$}ASPP} & $\mu$ & 0.901 & 0.610 & 1.195 & 0.831 & \textbf{0.715} & 0.759 & 2.420 & 0.720 \\
    & & $\sigma$ & 0.001 & 0.004 & 0.013 & 0.004 & 0.003 & 0.004 & 0.015 & 0.022 \\
    \midrule
    \multirow{4}{*}{\scalebox{1.10}{OSIE}} & \multirow{2}{*}{\raisebox{1.0pt}{$\oplus$}ASPP} & $\mu$ & \hspace{0.4em}\textbf{0.918}* & \hspace{0.4em}\textbf{0.648}* & \hspace{0.4em}\textbf{1.647}* & \hspace{0.4em}\textbf{0.816}* & \hspace{0.4em}\textbf{0.788}* & \hspace{0.4em}\textbf{0.808}* & \textbf{3.010} & \textbf{0.749} \\
    & & $\sigma$ & 0.001 & 0.002 & 0.017 & 0.004 & 0.004 & 0.002 & 0.010 & 0.031 \\
    \addlinespace[0.15em]
    & \multirow{2}{*}{\raisebox{1.0pt}{$\ominus$}ASPP} & $\mu$ & 0.916 & 0.641 & 1.733 & 0.808 & 0.781 & 0.804 & 3.000 & 0.767 \\
    & & $\sigma$ & 0.001 & 0.002 & 0.025 & 0.008 & 0.006 & 0.002 & 0.010 & 0.046 \\
    \bottomrule
\end{tabularx}}
\caption{A summary of the quantitative results for the models with $\oplus$ and without $\ominus$ an ASPP module. The evaluation was carried out on five eye tracking datasets respectively. Each network was independently trained 10 times resulting in a distribution of values characterized by the mean $\mu$ and standard deviation $\sigma$. The star * denotes a significant increase of performance between the two conditions according to a one sided paired t-test. Arrows indicate whether the metrics assess similarity \protect\raisebox{1.0pt}{\protect\scalebox{0.70}{$\uparrow$}} or dissimilarity \protect\raisebox{1.0pt}{\protect\scalebox{0.70}{$\downarrow$}} between predictions and targets. The best results are marked in bold.}
\label{tab:tab6}
\end{table}

\newpage

To quantify the contribution of multi-scale contextual information to the overall performance, we conducted a model ablation analysis. A baseline architecture without the ASPP module was constructed by replacing the five parallel convolutional layers with a single $3\times3$ convolutional operation that resulted in 1,280 activation maps. This representation was then forwarded to a $1\times1$ convolutional layer with 256 channels. While the total number of feature maps stayed constant, the amount of trainable parameters increased in this ablation setting. Table~\ref{tab:tab6} summarizes the results according to validation instances of five eye tracking datasets for the model with and without an ASPP module. It can be seen that our multi-scale architecture reached significantly higher performance (one tailed paired t-test) on most metrics and is therefore able to leverage the information captured by convolutional layers with different receptive field sizes. An ablation analysis of the multi-level component adapted from~\citet{cornia2016deep} can be viewed in the~\ref{app:app1}.

\begin{figure}[t!]
\centering\includegraphics[width=1.0\linewidth]{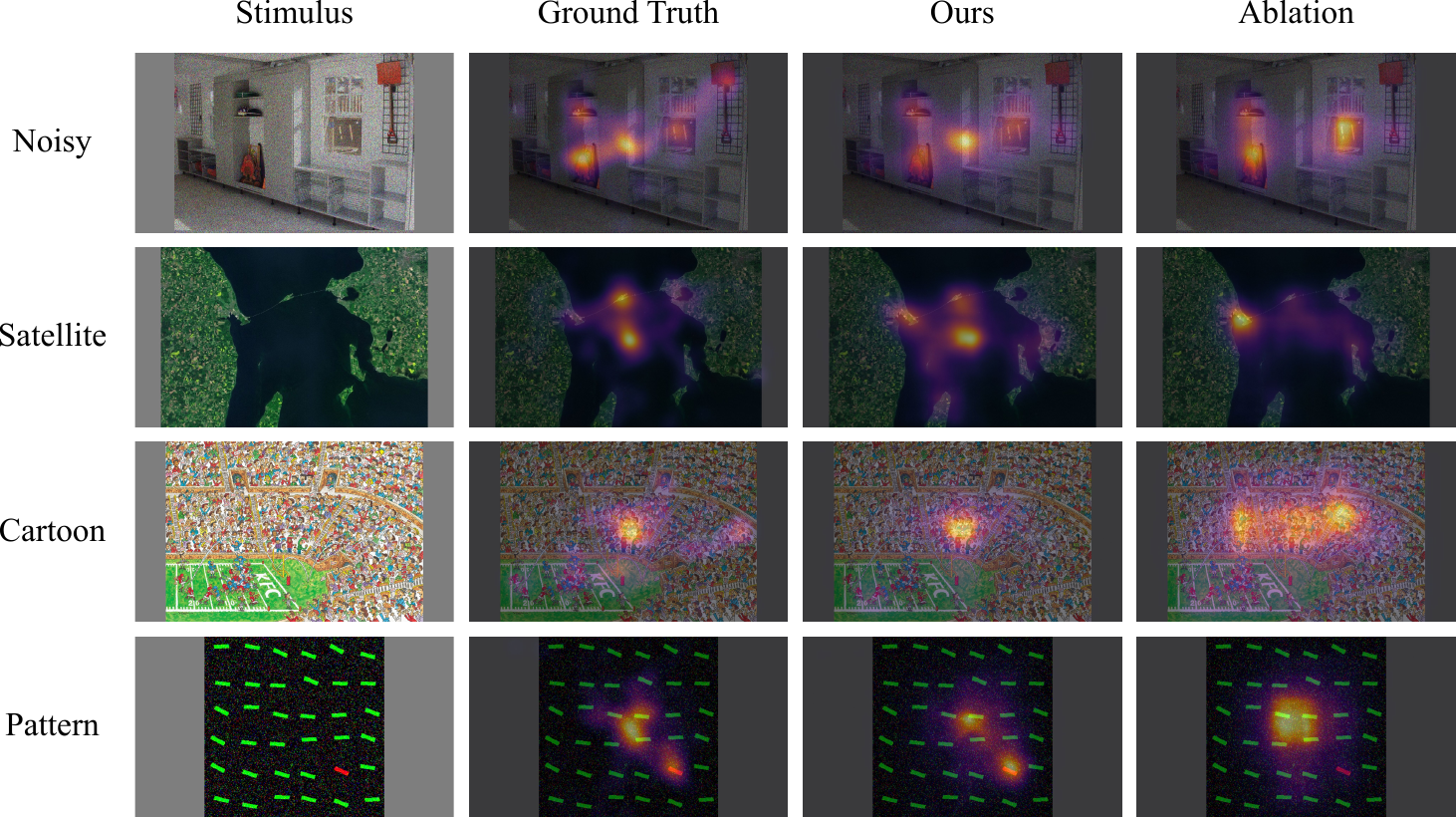}
\caption{A visualization of four example images from the CAT2000 validation set with the corresponding fixation heat maps, our best model predictions, and estimated maps based on the ablated network. The qualitative results indicate that multi-scale information augmented with global context enables a more accurate estimation of salient image regions.}
\label{fig:fig3}
\end{figure}

\begin{table}[t!]
\centering
\scalebox{0.76}{
\begin{tabularx}{17.6cm}{P{4.12cm} Y Y Y Y Y Y Y Y}
    \toprule
    & \textit{\scalebox{0.9}{AUC-J}} \raisebox{1.3pt}{\scalebox{0.64}{$\uparrow$}} & 
    \textit{\scalebox{0.9}{SIM}} \raisebox{1.3pt}{\scalebox{0.64}{$\uparrow$}} & 
    \textit{\scalebox{0.9}{EMD}} \raisebox{1.3pt}{\scalebox{0.64}{$\downarrow$}} & 
    \textit{\scalebox{0.9}{AUC-B}} \raisebox{1.3pt}{\scalebox{0.64}{$\uparrow$}} & 
    \textit{\scalebox{0.9}{sAUC}} \raisebox{1.3pt}{\scalebox{0.64}{$\uparrow$}} & 
    \textit{\scalebox{0.9}{CC}} \raisebox{1.3pt}{\scalebox{0.64}{$\uparrow$}} & 
    \textit{\scalebox{0.9}{NSS}} \raisebox{1.3pt}{\scalebox{0.64}{$\uparrow$}} & 
    \textit{\scalebox{0.9}{KLD}} \raisebox{1.3pt}{\scalebox{0.64}{$\downarrow$}} \\
    \midrule
    Noisy & \hspace{-0.4em}\raisebox{1.0pt}{\color{tab_green}\scalebox{0.85}{$+$}}\color{tab_green}0.010 & \hspace{-0.4em}\raisebox{1.0pt}{\color{tab_green}\scalebox{0.85}{$+$}}\color{tab_green}0.073 & \hspace{-0.4em}\raisebox{1.0pt}{\color{tab_green}\scalebox{0.85}{$-$}}\color{tab_green}0.506 & \hspace{-0.4em}\raisebox{1.0pt}{\color{tab_red}\scalebox{0.85}{$-$}}\color{tab_red}0.015 & \hspace{-0.4em}\raisebox{1.0pt}{\color{tab_red}\scalebox{0.85}{$-$}}\color{tab_red}0.009 & \hspace{-0.4em}\raisebox{1.0pt}{\color{tab_green}\scalebox{0.85}{$+$}}\color{tab_green}0.122 & \hspace{-0.4em}\raisebox{1.0pt}{\color{tab_green}\scalebox{0.85}{$+$}}\color{tab_green}0.395 & \hspace{-0.4em}\raisebox{1.0pt}{\color{tab_green}\scalebox{0.85}{$-$}}\color{tab_green}0.099 \\
    \addlinespace[0.10em]
    Satellite & \hspace{-0.4em}\raisebox{1.0pt}{\color{tab_green}\scalebox{0.85}{$+$}}\color{tab_green}0.015 & \hspace{-0.4em}\raisebox{1.0pt}{\color{tab_green}\scalebox{0.85}{$+$}}\color{tab_green}0.060 & \hspace{-0.4em}\raisebox{1.0pt}{\color{tab_green}\scalebox{0.85}{$-$}}\color{tab_green}0.663 & \hspace{-0.4em}\raisebox{1.0pt}{\color{tab_red}\scalebox{0.85}{$-$}}\color{tab_red}0.012 & \hspace{-0.4em}\raisebox{1.0pt}{\color{tab_red}\scalebox{0.85}{$-$}}\color{tab_red}0.007 & \hspace{-0.4em}\raisebox{1.0pt}{\color{tab_green}\scalebox{0.85}{$+$}}\color{tab_green}0.137 & \hspace{-0.4em}\raisebox{1.0pt}{\color{tab_green}\scalebox{0.85}{$+$}}\color{tab_green}0.362 & \hspace{-0.4em}\raisebox{1.0pt}{\color{tab_green}\scalebox{0.85}{$-$}}\color{tab_green}0.100 \\
    \addlinespace[0.10em]
    Cartoon & \hspace{-0.4em}\raisebox{1.0pt}{\color{tab_green}\scalebox{0.85}{$+$}}\color{tab_green}0.015 & \hspace{-0.4em}\raisebox{1.0pt}{\color{tab_green}\scalebox{0.85}{$+$}}\color{tab_green}0.066 & \hspace{-0.4em}\raisebox{1.0pt}{\color{tab_green}\scalebox{0.85}{$-$}}\color{tab_green}0.652 & \hspace{-0.4em}\raisebox{1.0pt}{\color{tab_red}\scalebox{0.85}{$-$}}\color{tab_red}0.010 & \hspace{-0.4em}\raisebox{1.0pt}{\color{tab_red}\scalebox{0.85}{$-$}}\color{tab_red}0.004 & \hspace{-0.4em}\raisebox{1.0pt}{\color{tab_green}\scalebox{0.85}{$+$}}\color{tab_green}0.125 & \hspace{-0.4em}\raisebox{1.0pt}{\color{tab_green}\scalebox{0.85}{$+$}}\color{tab_green}0.349 & \hspace{-0.4em}\raisebox{1.0pt}{\color{tab_green}\scalebox{0.85}{$-$}}\color{tab_green}0.121 \\
    \addlinespace[0.10em]
    Pattern & \hspace{-0.4em}\raisebox{1.0pt}{\color{tab_green}\scalebox{0.85}{$+$}}\color{tab_green}0.011 & \hspace{-0.4em}\raisebox{1.0pt}{\color{tab_green}\scalebox{0.85}{$+$}}\color{tab_green}0.050 & \hspace{-0.4em}\raisebox{1.0pt}{\color{tab_green}\scalebox{0.85}{$-$}}\color{tab_green}0.437 & \hspace{-0.4em}\raisebox{1.0pt}{\color{tab_red}\scalebox{0.85}{$-$}}\color{tab_red}0.003 & \hspace{-0.4em}\raisebox{1.0pt}{\color{tab_green}\scalebox{0.85}{$+$}}\color{tab_green}0.001 & \hspace{-0.4em}\raisebox{1.0pt}{\color{tab_green}\scalebox{0.85}{$+$}}\color{tab_green}0.078 & \hspace{-0.4em}\raisebox{1.0pt}{\color{tab_green}\scalebox{0.85}{$+$}}\color{tab_green}0.277 & \hspace{-0.4em}\raisebox{1.0pt}{\color{tab_green}\scalebox{0.85}{$-$}}\color{tab_green}0.065 \\
    \bottomrule
\end{tabularx}}
\caption{A list of the four image categories from the CAT2000 validation set that showed the largest average improvement by the ASPP architecture based on the cumulative rank across a subset of weakly correlated evaluation measures. Arrows indicate whether the metrics assess similarity \protect\raisebox{1.0pt}{\protect\scalebox{0.70}{$\uparrow$}} or dissimilarity \protect\raisebox{1.0pt}{\protect\scalebox{0.70}{$\downarrow$}} between predictions and targets. Results that improved on the respective metric are marked in green, whereas results that impaired performance are marked in red.}
\label{tab:tab7}
\end{table}

\newpage

The categorical organization of the CAT2000 database also allowed us to quantify the improvements by the ASPP module with respect to individual image classes. Table~\ref{tab:tab7} lists the four categories that benefited the most from multi-scale information across the subset of evaluation metrics on the validation set: \textit{Noisy}, \textit{Satellite}, \textit{Cartoon}, \textit{Pattern}. To understand the measured changes in predictive performance, it is instructive to inspect qualitative results of one representative example for each image category (see Figure~\ref{fig:fig3}). The visualizations demonstrate that large receptive fields allow the reweighting of relative importance assigned to image locations (\textit{Noisy}, \textit{Satellite}, \textit{Cartoon}), detection of a central fixation bias (\textit{Noisy}, \textit{Satellite}, \textit{Cartoon}), and allocation of saliency to a low-level color contrast that pops out from an array of distractors (\textit{Pattern}).

\begin{figure}[t!]
\centering\includegraphics[width=0.725\linewidth]{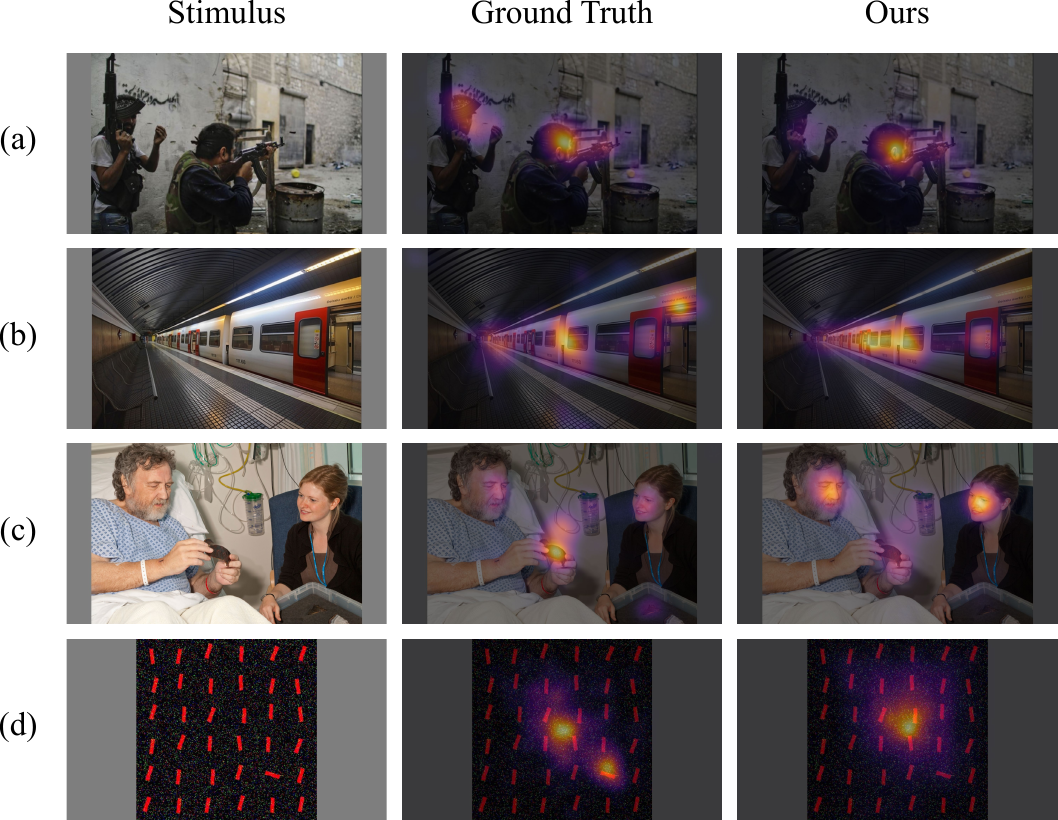}
\caption{A visualization of four example images from the CAT2000 validation set with the corresponding eye movement patterns and our model predictions. The stimuli demonstrate cases with a qualitative disagreement between the estimated saliency maps and ground truth data. Here, our model failed to capture an occluded face (a), small text (b), direction of gaze (c), and low-level feature contrast (d).}
\label{fig:fig4}
\end{figure}

\section{Discussion}

Our proposed encoder-decoder model clearly demonstrated competitive performance on two datasets towards visual saliency prediction. The ASPP module incorporated multi-scale information and global context based on semantic feature representations, which significantly improved the results both qualitatively and quantitatively on five eye tracking datasets. This suggests that convolutional layers with large receptive fields at different dilation factors can enable a more holistic estimation of salient image regions in complex scenes. Moreover, our approach is computationally lightweight compared to prior state-of-the-art approaches and could thus be implemented in (virtual) robotic systems that require computational efficiency. It also outperformed all other networks defined with a pre-trained VGG16 backbone as calculated by the cumulative rank on a subset of evaluation metrics to resolve some of the inconsistencies in ranking models by a single measure or a set of correlated ones~\cite{riche2013saliency,bylinskii2018different}.

Further improvements of benchmark results could potentially be achieved by a number of additions to the processing pipeline. Our model demonstrates a learned preference for predicting fixations in central regions of images, but we expect performance gains from modeling the central bias in scene viewing explicitly~\cite{kummerer2014deep,kummerer2016deepgaze,cornia2016deep,Cornia2018PredictingHE,kruthiventi2017deepfix}. Additionally, \citet{bylinskii2015saliency} summarized open problems for correctly assigning saliency in natural images, such as robustness in detecting semantic features, implied gaze and motion, and importance weighting of multiple salient regions. While the latter was addressed in this study, Figure~\ref{fig:fig4} indicates that the remaining obstacles still persist for our proposed model.

Overcoming these issues requires a higher-level scene understanding that models object interactions and predicts implicit gaze and motion cues from static images. Robust object recognition could however be achieved through more recent classification networks as feature extractors~\cite{oyama2018influence} at the cost of added computational complexity. However, this study does not investigate whether the benefits of the proposed modifications generalize to other pre-trained architectures. That would constitute an interesting avenue for future research. To detect salient items in search array stimuli (see Figure~\ref{fig:fig4}d), a mechanism that additionally captures low-level feature contrasts might explain the empirical data better. Besides architectural changes, data augmentation in the context of saliency prediction tasks demonstrated its efficiency to improve the robustness of deep neural networks according to~\citet{che2018invariance}. These authors stated that visual transformations such as mirroring or inversion revealed a low impact on human gaze during scene viewing and could hence form an addition to future work on saliency models.

\section*{Declaration of Competing Interest}

The authors declare that they have no known competing financial interests or personal relationships that could have appeared to influence the work reported in this paper.

\section*{Acknowledgement}

This study has received funding from the European Union's Horizon 2020 Framework Programme for Research and Innovation under the Specific Grant Agreement Nos. 720270 (Human Brain Project SGA1) and 785907 (Human Brain Project SGA2). Furthermore, we gratefully acknowledge the support of NVIDIA Corporation with the donation of a Titan X Pascal GPU used for this research.

\bibliographystyle{refstyle.bst}
\bibliography{references.bib}

\newpage
\appendix

\section{Feature Concatenation Ablation Analysis}
\label{app:app1}

\begin{table}[h!]
\centering
\scalebox{0.76}{
\begin{tabularx}{17.6cm}{P{2.25cm} P{1.7cm} P{0.05cm} Y Y Y Y Y Y Y Y}
    \toprule
    & & & \textit{\scalebox{0.9}{AUC-J}} \raisebox{1.3pt}{\scalebox{0.64}{$\uparrow$}} & 
    \textit{\scalebox{0.9}{SIM}} \raisebox{1.3pt}{\scalebox{0.64}{$\uparrow$}} & 
    \textit{\scalebox{0.9}{EMD}} \raisebox{1.3pt}{\scalebox{0.64}{$\downarrow$}} & 
    \textit{\scalebox{0.9}{AUC-B}} \raisebox{1.3pt}{\scalebox{0.64}{$\uparrow$}} & 
    \textit{\scalebox{0.9}{sAUC}} \raisebox{1.3pt}{\scalebox{0.64}{$\uparrow$}} & 
    \textit{\scalebox{0.9}{CC}} \raisebox{1.3pt}{\scalebox{0.64}{$\uparrow$}} & 
    \textit{\scalebox{0.9}{NSS}} \raisebox{1.3pt}{\scalebox{0.64}{$\uparrow$}} & 
    \textit{\scalebox{0.9}{KLD}} \raisebox{1.3pt}{\scalebox{0.64}{$\downarrow$}} \\
    \midrule
    \multirow{4}{*}{\scalebox{1.10}{MIT1003}} & \multirow{2}{*}{\raisebox{1.0pt}{$\oplus$}\hspace{0.05cm}CONCAT} & $\mu$ & \textbf{0.899} & \textbf{0.602} & \textbf{2.430} & 0.832 & 0.713 & \textbf{0.741} & \textbf{2.663} & 0.818 \\
    & & $\sigma$ & 0.001 & 0.004 & 0.033 & 0.005 & 0.003 & 0.003 & 0.012 & 0.051 \\
    \addlinespace[0.15em]
    & \multirow{2}{*}{\raisebox{1.0pt}{$\ominus$}\hspace{0.05cm}CONCAT} & $\mu$ & 0.898 & 0.599 & 2.445 & \textbf{0.837} & \textbf{0.715} & 0.740 & 2.649 & \textbf{0.794} \\
    & & $\sigma$ & 0.001 & 0.007 & 0.048 & 0.008 & 0.003 & 0.004 & 0.022 & 0.039 \\
    \midrule
    \multirow{4}{*}{\scalebox{1.10}{CAT2000}} & \multirow{2}{*}{\raisebox{1.0pt}{$\oplus$}\hspace{0.05cm}CONCAT} & $\mu$ & \textbf{0.882} & \textbf{0.734} & 2.553 & 0.812 & 0.582 & 0.854 & \textbf{2.359} & \textbf{0.430} \\
    & & $\sigma$ & 0.000 & 0.002 & 0.025 & 0.005 & 0.003 & 0.003 & 0.007 & 0.010 \\
    \addlinespace[0.15em]
    & \multirow{2}{*}{\raisebox{1.0pt}{$\ominus$}\hspace{0.05cm}CONCAT} & $\mu$ & 0.881 & \textbf{0.734} & \textbf{2.545} & \textbf{0.814} & \textbf{0.586} & \textbf{0.855} & 2.354 & 0.436 \\
    & & $\sigma$ & 0.000 & 0.003 & 0.032 & 0.003 & 0.003 & 0.002 & 0.012 & 0.013 \\
    \midrule
    \multirow{4}{*}{\scalebox{1.10}{DUT-OMRON}} & \multirow{2}{*}{\raisebox{1.0pt}{$\oplus$}\hspace{0.05cm}CONCAT} & $\mu$ & \hspace{0.4em}\textbf{0.921}* & \hspace{0.4em}\textbf{0.649}* & \hspace{0.4em}\textbf{1.155}* & \textbf{0.864} & \textbf{0.775} & \hspace{0.4em}\textbf{0.776}* & \hspace{0.4em}\textbf{2.873}* & \hspace{0.4em}\textbf{0.634}* \\
    & & $\sigma$ & 0.001 & 0.002 & 0.016 & 0.005 & 0.003 & 0.002 & 0.008 & 0.023 \\
    \addlinespace[0.15em]
    & \multirow{2}{*}{\raisebox{1.0pt}{$\ominus$}\hspace{0.05cm}CONCAT} & $\mu$ & 0.919 & 0.641 & 1.191 & 0.861 & 0.773 & 0.766 & 2.825 & 0.656 \\
    & & $\sigma$ & 0.001 & 0.003 & 0.018 & 0.005 & 0.003 & 0.003 & 0.009 & 0.042 \\
    \midrule
    \multirow{4}{*}{\scalebox{1.10}{PASCAL-S}} & \multirow{2}{*}{\raisebox{1.0pt}{$\oplus$}\hspace{0.05cm}CONCAT} & $\mu$ & \hspace{0.4em}\textbf{0.914}* & \hspace{0.4em}\textbf{0.667}* & \hspace{0.4em}\textbf{1.015}* & 0.853 & 0.701 & \hspace{0.4em}\textbf{0.818}* & \hspace{0.4em}\textbf{2.610}* & \textbf{0.645} \\
    & & $\sigma$ & 0.000 & 0.003 & 0.010 & 0.004 & 0.003 & 0.002 & 0.008 & 0.044 \\
    \addlinespace[0.15em]
    & \multirow{2}{*}{\raisebox{1.0pt}{$\ominus$}\hspace{0.05cm}CONCAT} & $\mu$ & 0.907 & 0.636 & 1.130 & \textbf{0.855} & \textbf{0.711} & 0.791 & 2.494 & 0.649 \\
    & & $\sigma$ & 0.001 & 0.006 & 0.019 & 0.004 & 0.005 & 0.005 & 0.017 & 0.031 \\
    \midrule
    \multirow{4}{*}{\scalebox{1.10}{OSIE}} & \multirow{2}{*}{\raisebox{1.0pt}{$\oplus$}\hspace{0.05cm}CONCAT} & $\mu$ & \hspace{0.4em}\textbf{0.918}* & \hspace{0.4em}\textbf{0.648}* & \hspace{0.4em}\textbf{1.647}* & 0.816 & \textbf{0.788} & \hspace{0.4em}\textbf{0.808}* & \hspace{0.4em}\textbf{3.010}* & 0.749 \\
    & & $\sigma$ & 0.001 & 0.002 & 0.017 & 0.004 & 0.004 & 0.002 & 0.010 & 0.031 \\
    \addlinespace[0.15em]
    & \multirow{2}{*}{\raisebox{1.0pt}{$\ominus$}\hspace{0.05cm}CONCAT} & $\mu$ & 0.908 & 0.605 & 1.932 & \textbf{0.821} & \textbf{0.788} & 0.760 & 2.774 & \textbf{0.740} \\
    & & $\sigma$ & 0.001 & 0.005 & 0.028 & 0.009 & 0.007 & 0.005 & 0.027 & 0.039 \\
    \bottomrule
\end{tabularx}}
\caption{A summary of the quantitative results for the models with $\oplus$ and without $\ominus$ the concatenation of encoder features. The evaluation was carried out on five eye tracking datasets respectively. Each network was independently trained 10 times resulting in a distribution of values characterized by the mean $\mu$ and standard deviation $\sigma$. The star * denotes a significant increase of performance between the two conditions according to a one sided paired t-test. Arrows indicate whether the metrics assess similarity \protect\raisebox{1.0pt}{\protect\scalebox{0.70}{$\uparrow$}} or dissimilarity \protect\raisebox{1.0pt}{\protect\scalebox{0.70}{$\downarrow$}} between predictions and targets. The best results are marked in bold.}
\label{tab:tab8}
\end{table}
\begin{table}[h!]
\centering
\scalebox{0.76}{
\begin{tabularx}{17.6cm}{P{4.12cm} Y Y Y Y Y Y Y Y}
    \toprule
    & \textit{\scalebox{0.9}{AUC-J}} \raisebox{1.3pt}{\scalebox{0.64}{$\uparrow$}} & 
    \textit{\scalebox{0.9}{SIM}} \raisebox{1.3pt}{\scalebox{0.64}{$\uparrow$}} & 
    \textit{\scalebox{0.9}{EMD}} \raisebox{1.3pt}{\scalebox{0.64}{$\downarrow$}} & 
    \textit{\scalebox{0.9}{AUC-B}} \raisebox{1.3pt}{\scalebox{0.64}{$\uparrow$}} & 
    \textit{\scalebox{0.9}{sAUC}} \raisebox{1.3pt}{\scalebox{0.64}{$\uparrow$}} & 
    \textit{\scalebox{0.9}{CC}} \raisebox{1.3pt}{\scalebox{0.64}{$\uparrow$}} & 
    \textit{\scalebox{0.9}{NSS}} \raisebox{1.3pt}{\scalebox{0.64}{$\uparrow$}} & 
    \textit{\scalebox{0.9}{KLD}} \raisebox{1.3pt}{\scalebox{0.64}{$\downarrow$}} \\
    \midrule
    Action & \hspace{-0.4em}\raisebox{1.0pt}{\color{tab_green}\scalebox{0.85}{$+$}}\color{tab_green}0.001 & \hspace{-0.4em}\raisebox{1.0pt}{\color{tab_green}\scalebox{0.85}{$+$}}\color{tab_green}0.007 & \hspace{-0.4em}\raisebox{1.0pt}{\color{tab_green}\scalebox{0.85}{$-$}}\color{tab_green}0.062 & \hspace{-0.4em}\raisebox{1.0pt}{\color{tab_green}\scalebox{0.85}{$+$}}\color{tab_green}0.001 & \hspace{-0.4em}\raisebox{0.4pt}{\scalebox{0.85}{$\pm$}}0.000 & \hspace{-0.4em}\raisebox{1.0pt}{\color{tab_green}\scalebox{0.85}{$+$}}\color{tab_green}0.010 & \hspace{-0.4em}\raisebox{1.0pt}{\color{tab_green}\scalebox{0.85}{$+$}}\color{tab_green}0.025 & \hspace{-0.4em}\raisebox{1.0pt}{\color{tab_green}\scalebox{0.85}{$-$}}\color{tab_green}0.020 \\
    \addlinespace[0.10em]
    Social & \hspace{-0.4em}\raisebox{1.0pt}{\color{tab_green}\scalebox{0.85}{$+$}}\color{tab_green}0.004 & \hspace{-0.4em}\raisebox{1.0pt}{\color{tab_green}\scalebox{0.85}{$+$}}\color{tab_green}0.003 & \hspace{-0.4em}\raisebox{1.0pt}{\color{tab_green}\scalebox{0.85}{$-$}}\color{tab_green}0.064 & \hspace{-0.4em}\raisebox{1.0pt}{\color{tab_green}\scalebox{0.85}{$+$}}\color{tab_green}0.002 & \hspace{-0.4em}\raisebox{1.0pt}{\color{tab_green}\scalebox{0.85}{$+$}}\color{tab_green}0.002 & \hspace{-0.4em}\raisebox{1.0pt}{\color{tab_green}\scalebox{0.85}{$+$}}\color{tab_green}0.007 & \hspace{-0.4em}\raisebox{1.0pt}{\color{tab_green}\scalebox{0.85}{$+$}}\color{tab_green}0.025 & \hspace{-0.4em}\raisebox{1.0pt}{\color{tab_green}\scalebox{0.85}{$-$}}\color{tab_green}0.037 \\
    \midrule
    Fractal & \hspace{-0.4em}\raisebox{1.0pt}{\color{tab_green}\scalebox{0.85}{$+$}}\color{tab_green}0.001 & \hspace{-0.4em}\raisebox{1.0pt}{\color{tab_red}\scalebox{0.85}{$-$}}\color{tab_red}0.001 & \hspace{-0.4em}\raisebox{1.0pt}{\color{tab_red}\scalebox{0.85}{$+$}}\color{tab_red}0.034 & \hspace{-0.4em}\raisebox{1.0pt}{\color{tab_red}\scalebox{0.85}{$-$}}\color{tab_red}0.017 & \hspace{-0.4em}\raisebox{1.0pt}{\color{tab_red}\scalebox{0.85}{$-$}}\color{tab_red}0.004 & \hspace{-0.4em}\raisebox{0.4pt}{\scalebox{0.85}{$\pm$}}0.000 & \hspace{-0.4em}\raisebox{1.0pt}{\color{tab_green}\scalebox{0.85}{$+$}}\color{tab_green}0.018 & \hspace{-0.4em}\raisebox{1.0pt}{\color{tab_red}\scalebox{0.85}{$+$}}\color{tab_red}0.018 \\
    \addlinespace[0.10em]
    Pattern & \hspace{-0.4em}\raisebox{0.4pt}{\scalebox{0.85}{$\pm$}}0.000 & \hspace{-0.4em}\raisebox{1.0pt}{\color{tab_green}\scalebox{0.85}{$+$}}\color{tab_green}0.006 & \hspace{-0.4em}\raisebox{1.0pt}{\color{tab_green}\scalebox{0.85}{$-$}}\color{tab_green}0.051 & \hspace{-0.4em}\raisebox{1.0pt}{\color{tab_red}\scalebox{0.85}{$-$}}\color{tab_red}0.005 & \hspace{-0.4em}\raisebox{1.0pt}{\color{tab_red}\scalebox{0.85}{$-$}}\color{tab_red}0.004 & \hspace{-0.4em}\raisebox{0.4pt}{\scalebox{0.85}{$\pm$}}0.000 & \hspace{-0.4em}\raisebox{1.0pt}{\color{tab_red}\scalebox{0.85}{$-$}}\color{tab_red}0.005 & \hspace{-0.4em}\raisebox{1.0pt}{\color{tab_red}\scalebox{0.85}{$+$}}\color{tab_red}0.016 \\
    \bottomrule
\end{tabularx}}
\caption{A list of the image categories from the CAT2000 validation set that either showed the largest average improvement (first two entries) or impairment (last two entries) by the multi-level design based on the cumulative rank across a subset of weakly correlated evaluation measures. Arrows indicate whether the metrics assess similarity \protect\raisebox{1.0pt}{\protect\scalebox{0.70}{$\uparrow$}} or dissimilarity~\protect\raisebox{1.0pt}{\protect\scalebox{0.70}{$\downarrow$}} between predictions and targets. Results that improved on the respective metric are marked in green, whereas results that impaired performance are marked in red.}
\label{tab:tab9}
\end{table}

In this experimental setting, we removed the concatenation operation from the network architecture and compared the model performance of the ablated version to the one including a multi-level design (see Table~\ref{tab:tab8}). While models trained on the CAT2000 dataset did not consistently benefit from the aggregation of features at different stages of the encoder, all other cases demonstrated a mostly significant improvement according to the majority of metric scores. Table~\ref{tab:tab9} indicates that predictions on natural image categories (\textit{Action}, \textit{Social}) leveraged the multi-level information for better performance, whereas adverse results were achieved on artificial and simplified stimuli (\textit{Fractal}, \textit{Pattern}). In conclusion, the feature concatenation design might only be recommendable for training models on datasets that mostly consist of complex natural images, such as MIT1003, DUT-OMRON, PASCAL-S, or OSIE.

\end{document}